\documentclass[letterpaper, 10 pt, conference]{ieeeconf}

\IEEEoverridecommandlockouts                              

\usepackage{algorithm,algpseudocode}
\usepackage{amsmath, graphicx } 
\usepackage{balance}
\usepackage{subcaption} 
\usepackage{amssymb,amsbsy}
\usepackage{breqn,bbm,xcolor}
\usepackage{multirow}
\usepackage[symbol]{footmisc}

\title{\LARGE \bf
SuPer Deep: A Surgical Perception Framework for Robotic Tissue Manipulation using Deep Learning for Feature Extraction}

\author{Jingpei Lu$^{1*}$, Ambareesh Jayakumari$^{1*}$, Florian Richter$^1$ \IEEEmembership{Student Member, IEEE},\\ Yang Li$^2$ and Michael C. Yip$^{1}$ \IEEEmembership{Senior Member, IEEE}
\thanks{$^{*}$Equal contributions. }%
\thanks{$^{1}$Jingpei Lu, Ambareesh Jayakumari, Florian Richter and Michael C. Yip are with the Department of Electrical and Computer Engineering, University of California San Diego, La Jolla, CA 92093 USA.
{\tt\small\{jil360, asreekum, frichter, yip\}@ucsd.edu}}%
\thanks{$^{2}$Yang Li is with the School of Computer Science and Technology, East China Normal University, Shanghai, 200062 China.
{\tt\small{yli@cs.ecnu.edu.cn}}}%
}

\begin{document}

\maketitle
\thispagestyle{empty}
\pagestyle{empty}

\begin{abstract}
Robotic automation in surgery requires precise tracking of surgical tools and mapping of deformable tissue.
Previous works on surgical perception frameworks require significant effort in developing features for surgical tool and tissue tracking. 
In this work, we overcome the challenge by exploiting deep learning methods for surgical perception. We integrated deep neural networks, capable of efficient feature extraction, into the tissue tracking and surgical tool tracking processes. By leveraging transfer learning, the deep-learning-based approach requires minimal training data and reduced feature engineering efforts to fully perceive a surgical scene.
The framework was tested on three publicly available datasets, which use the da Vinci\textregistered{} Surgical System, for comprehensive analysis. 
Experimental results show that our framework achieves state-of-the-art tracking performance in a surgical environment by utilizing deep learning for feature extraction.
\end{abstract}

\section{INTRODUCTION}

In the field of health care and surgery, automation is on the horizon due to advancements in robotics. Improved patient outcomes are being achieved through increased precision in tissue manipulation and the development of minimally invasive robotics \cite{ballantyne2003vinci}. 
One avenue of research in automation using these platforms is through the advancements of control algorithms to move towards autonomy \cite{yipDasJournal, dvrl}. 
These algorithms typically aim to automate specific surgical subtasks such as suturing \cite{suturing}, cutting \cite{surgical_cutting_rl}, and multilateral debridement \cite{debridement_removal}. 
Another development is assistance to the teleoperating surgeon in real-time through virtual fixtures to avoid critical areas \cite{virtual_fixtures},  augmented reality indicators \cite{qian2019review}, and motion scaling for finer control near tissue \cite{zhang2018self}. 

To utilize these automation endeavors in a real surgical scene, accurate perception of the environment and the agents is essential. 
There are two major challenges: tracking of the surgical tool to control and localize it in the camera frame, and tracking of the deformable environment for the surgical tool to plan and interact with. 
While these two problems have been solved outside of surgical robotics \cite{li2019robust,gao18surfelwarp}, the domain-specific challenges are the narrow field of view endoscopes, poor lighting conditions, and the requirement of very high accuracy \cite{bouget2017vision}.

%

The surgical tool tracking community has largely focused on developing feature detection algorithms to update the pose of the surgical tool \cite{bouget2017vision}. 
The algorithms need to be robust to the poor lighting conditions and the highly reflective tool surfaces.
Examples of recent work include using the Canny-edge detector for silhouette extraction \cite{Hao}, online template matching \cite{EKF}, and classified features using classical image features such as the spatial derivatives \cite{Reiter}, \cite{Reiter_tracking}.
Deep neural networks have also achieved promising results in feature tracking for surgical tools \cite{Kurmann_2017}, \cite{colleoni_tool}, but utilizing them for full 3D pose estimation still remains unexamined.

Simultaneously, efforts in tissue tracking have focused mainly on adaptions of 3D reconstruction techniques such as SurfelWarp for deformable tracking \cite{gao18surfelwarp}. The lack of directly measurable depth information in endoscopes is a  significant challenge in the adaptation.
Hence, the common approach is to work with stereoscopic endoscopes and use stereo reconstruction techniques such as Efficient Large-Scale Stereo Matching (ELAS) to generate depth images \cite{stereo_matching}. 
From this depth estimation, deformable tracking techniques can be applied \cite{song2017dynamicTissue, song2018mis}. 
Other tissue tracking techniques include tracking key-point features and registration \cite{yip2012tissue} and dense SLAM methods, which use image features to localize the endoscope \cite{mahmoud2018live, marmol2019dense}.

\label{section:methodology}
\begin{figure*}[ht!]
\vspace{2mm}
    \centering
    \includegraphics[width=1.0\linewidth]{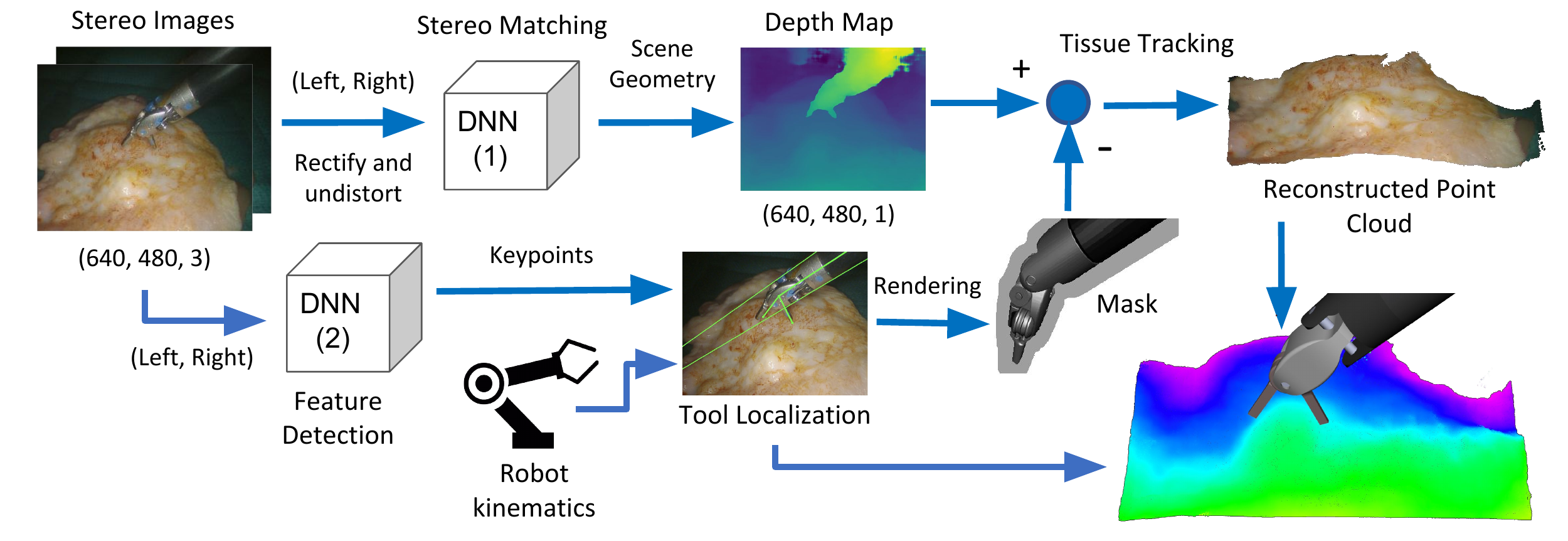}
    \caption{The complete workflow of the proposed SuPer Deep framework. The DNN(1) is used to match the features from stereo images to generate the depth map for deformable tissue tracking. The DNN(2) is employed to detect the point features for estimating the pose of the surgical tool.
    The entire surgical scene is reconstructed by fusing the depth maps into the deformable tissue model and localizing the tools in the camera frame using the tool tracker.}
    \label{fig:blockdiagram}
\end{figure*}

A common theme across these two challenges is the need for high-quality image features. Surgical tool tracking mainly focuses on developing detectors for tool features, and recent works in tissue tracking have highlighted depth reconstruction from stereo matching as the most significant bottleneck \cite{song2017dynamicTissue, li2019super}.
Deep learning has the advantage of learning features, which will eliminate the need for feature engineering. 
However, deep learning previously has not been a front runner in surgical perception due to the lack of large quantities of high-quality
medical and surgical data \cite{kassahun2016surgical}.

In this paper, we use state-of-the-art deep neural networks (DNNs) that require minimal training data to explore its application in surgical perception.
Our contributions can be summarized as follows:
\begin{enumerate}
    \item Using deep learning for high quality and robust surgical tool feature extraction,
    \item Investigative study of popular deep learning and traditional stereo matching algorithms to improve deformable tissue tracking,
    \item Complete integration of deep neural networks into the previously developed Surgical Perception (SuPer) framework  \cite{li2019super} to fully perceive the entire surgical scene - \textit{SuPer Deep}. 
\end{enumerate}
 
Experiments were ran on a tissue manipulation dataset we previously released \cite{li2019super}, and two other publicly available datasets collected from the da Vinci\textregistered{} Surgical System, to evaluate the tool tracking and tissue tracking performances individually. 
SuPer Deep framework advances on SuPer \cite{li2019super}, its predecessor, by eliminating the requirement for painted markers and the key-points association process in tool tracking. Moreover, our framework generates more realistic tissue reconstruction by improving the depth estimation. Finally, our work includes the first comparative study on deformable tissue tracking, whereas previous works have been hampered by a paucity of standard benchmarking datasets for surgical perception.

\section{METHODOLOGY}

\subsection{The Surgical Perception Framework}
\label{sec:super_framework}

Our surgical perception framework, as shown in Fig. \ref{fig:blockdiagram}, consists of a deformable tissue tracker and a surgical tool tracker to perceive the entire surgical scene, including the the deforming environment and robotic agent. Two deep neural networks are embedded into our framework for specific feature extractions: DNN(1) finds and matches features from stereo images to generate a depth map for tissue tracking, and DNN(2) extracts point features for surgical tool tracking.


To track the tissue deformation, we update tissue model by fusing the estimated depth maps using the previously developed model-free tissue tracker \cite{li2019super}.
The pose of the surgical tool is estimated using a model-based tracker that utilizes a kinematic prior and fusing the encoder readings with the 2D detection from the images, which will be described in Section \ref{sec:tool_tracking}.
In order to separate the tools from the deformable tissue model, a mask of the surgical tool is generated by rendering the 3D CAD model into the endoscopic camera view and is removed from the depth map fed to the deformable tissue tracker.
Finally, we combine the tissue point cloud and surgical tools into the camera frame to reconstruct the entire surgical scene.

\subsection{Surgical Tool Tracking}
\label{sec:tool_tracking}

\begin{figure}[t]
\vspace{2mm}
\centering
\begin{subfigure}{0.23\textwidth}
\includegraphics[width=1\textwidth]{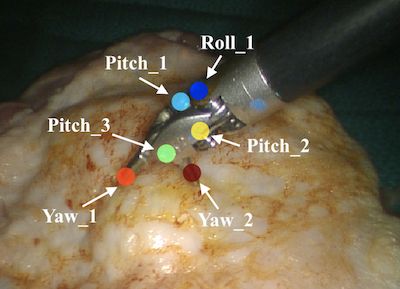}
\vspace{-0.12in}
\end{subfigure}
\begin{subfigure}{0.23\textwidth}
\includegraphics[width=1\textwidth]{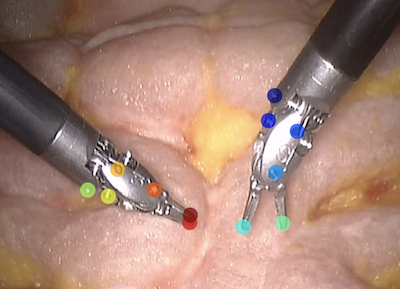}
\vspace{-0.12in}
\end{subfigure}

\caption{Illustration of the point features to detect on surgical tools. The left figure shows the features detected on a single instrument, and the right figure shows the features detected and associated appropriately with two instruments.  }
\label{fig:label_tool}
\vspace{-0.1in}
\end{figure}

To localize the surgical tools on image frames, we use DeepLabCut \cite{mathis2018deeplabcut}, which employs DeeperCut \cite{insafutdinov2016eccv} as the backbone for point feature detection.
The DNN consists of variations of Deep Residual Neural Networks (ResNet) \cite{resnet} for feature extraction and deconvolutional layers to up-sample the feature maps and produce spatial probability densities. 
The output estimation for each point feature is represented as a tuple $(\mathbf{h}^i, \rho^i)$, where $\mathbf{h}^i \in \mathbb{R}^2$ is the image coordinate of the $i$-th feature and $\rho^i \in \mathbb{R}$ is the corresponding confidence score. 
The DNN was fine-tuned with few training samples to adapt to surgical tool tracking by minimizing the cross-entropy loss.
The samples were hand-labeled using the open-source DLC toolbox \cite{dlc_tool}.
Fig. \ref{fig:label_tool} shows examples of point features that were detected on surgical instruments. 

To estimate the pose of the surgical tool in 3D space, the 2D detections are combined with the encoder readings from the surgical robot, and a particle filter is applied for estimation. 
Previous work has shown low-profile, cable driven surgical robots \cite{raven, dvrk} have inaccurate in joint angles \cite{seita2018fast} and challenges calibrating the transform between the camera and robotic base, also known as hand-eye \cite{calibration_dvrk_paper}.
To localize the surgical tool and overcome these uncertainties, we utilize our previously developed formulation and track the Lumped Error. \cite{richter2021robotic}.
Let the Lumped Error be defined as $\mathbf{T}^{b-}_b(\boldsymbol{\omega}, \mathbf{b}) \in SE(3)$, which is parameterized by an axis-angle vector $\boldsymbol{\omega} \in \mathbb{R}^3$, and a translational vector $\mathbf{b} \in \mathbb{R}^3$. 
The Lumped Error compensates for both error in joint angles and hand-eye, for more details refer to our previous work \cite{richter2021robotic}.
To estimate $\boldsymbol{\omega}$ and $\mathbf{b}$ at time $t+1$ given time $t$, a zero-mean Gaussian noise is assumed to model the uncertainty. 
Therefore, the motion model for the particle filter is defined as:
\begin{equation}
    [\boldsymbol{\omega}_{t+1|t}, \mathbf{b}_{t+1|t}]^T \sim \mathcal{N}([\boldsymbol{\omega}_{t|t}, \mathbf{b}_{t|t}]^T, \mathbf{\Sigma_{\boldsymbol{\omega},b}} )
\end{equation}
where $\mathbf{\Sigma_{\boldsymbol{\omega},b}}$ is the covariance matrix.

Using this Lumped Error formulation, a detected feature point can be projected to the image plane by being first transformed by the kinematic chain and then the corrected hand-eye.
More specifically, feature point $i$ on the $j_i$-th link $\mathbf{p}^{j_i} \in \mathbb{R}^3$ is projected onto the image plane by:
\begin{equation}
    \overline{\mathbf{m}}^i(\boldsymbol{\omega}, \mathbf{b}) = \frac{1}{s} \mathbf{K} \mathbf{T}^c_{b-} \mathbf{T}^{b-}_b (\boldsymbol{\omega}, \mathbf{b}) \prod \limits_{i=1}^{j_i} \mathbf{T}_i^{i-1} (\theta_i) \overline{\mathbf{p}}^{j_i}
    \label{projection_equation}
\end{equation}
where $\mathbf{T}_i^{i-1}(\theta_i) \in SE(3)$ is the $i$-th joint transform generated by joint angle $\theta_i$, $\mathbf{T}^c_{b-}$ is the initial hand-eye transform from the set-up joints or calibration, $\mathbf{K}$ is the intrinsic camera matrix, and $s$ is the scaling factor that constraints the point on the image plane.
Note that $\overline{\cdot}$ denotes the homogeneous representation of a point (e.g. $\overline{\mathbf{p}} = [\mathbf{p}, 1]^T$).

From here, an observation model can be defined by relating the predicted feature location with the detected features. 
Given a list of $N$ observations, $(\mathbf{h}_{t+1}, \boldsymbol{\rho}_{t+1})$, the observation model is defined to be:
\begin{multline}
    P( \mathbf{h}_{t+1}, \boldsymbol{\rho}_{t+1}  | \boldsymbol{\omega}_{t+1|t}, \mathbf{b}_{t+1|t})  \\ \propto \sum_{ i = 1}^{N} \rho^i_t e^{-\gamma ||\mathbf{h}^i_{t+1} - \mathbf{m}^i(\boldsymbol{\omega}_{t+1|t}, \mathbf{b}_{t+1|t}) ||^2}
    \label{observation_model_equation}
\end{multline}
where $\gamma$ is a tuning parameter.

Since the deep neural network performs end-to-end 2D feature detection, the only parameters that need to be tuned for tool tracking are $\gamma$ and $\mathbf{\Sigma_{\boldsymbol{\omega},b}}$. The direct association of the key-point features from the neural network also eliminates the need for explicit data association - a significant improvement for tool tracking over the previous method \cite{li2019super}.
Furthermore, the feature detection from the deep neural network does not rely on the tracked states, which is a common technique in surgical tool tracking \cite{EKF} and can lead to detrimental results when the tracking begins to fail.

\subsection{Depth Estimation for Deformable Tissue Tracking}

\label{sec:tissue_tracking}
Deformable tissue tracking relies heavily on the quality of depth estimation, as the deformable tracker uses the depth maps as the observation \cite{song2017dynamicTissue}, \cite{li2019super}.
To estimate the depth, a stereo matching algorithm is used to compute the disparity, and then inverted to obtain pixel-wise depth.

Traditional stereo matching algorithms, like \cite{stereo_matching} and \cite{SGBM}, typically take a pair of rectified stereo images $I_l$ and $I_r$ as input, and estimate the disparity by matching image patches or features between $I_l$ and $I_r$. Due to the complexity of the surgical setting, the image quality is not the sharpest, which makes finding pixel-level correspondence extremely challenging. 
Deep-learning-based algorithms, such as \cite{Kendall_2017}, \cite{Chang_2018}, and \cite{Zhang2019GANet}, use a weight-sharing feature extractor to obtain feature maps $F_l$ and $F_r$ from $I_l$ and $I_r$, respectively. Then a 4D matching cost volume $C_d$ is formed by concatenating the $F_l$ and $F_r$, such that $C_d(i,j,d,:)$ is the concatenation of $F_l(i,j)$ and $F_r(i,j+d)$, where $(i,j)$ is the pixel location and $d$ is the disparity. The cost volume is regularized using 3D convolutional layers and reducing the dimension of the 4-th channel to 1. Finally, the resulting 3D tensor $S_d$ is used to estimate the disparity for each pixel as
\begin{equation}
    \hat{d}(i,j) = \sum_{d=0}^{D_{max}}  \sigma(-S_d(i,j,d)) d
\end{equation}
where $D_{max}$ is the max disparity and $\sigma(\cdot)$ denotes the softmax function. An investigation between these stereo matching algorithms will be presented in the Section \ref{section:experiment}.
After estimating the disparity $\hat{d}$, the depth value $z$ is obtained by the following transform
\begin{equation}
    z(i,j) = \frac{bf}{\hat{d}(i,j)}
    \label{triangulation}
\end{equation}
where $b$ is the horizontal offset (i.e., baseline) between the two cameras, and $f$ is the focal length, which can be obtained from camera calibration. 

Various stereo-matching algorithms can be substituted in for DNN(1) of our framework shown in Fig. 1. To find the best one, we investigated several stereo matching algorithms by combining with the previously developed tissue tracker \cite{li2019super} for deformable tissue tracking. The state-of-the-art algorithm, Guided Aggregation Network (GA-Net) \cite{Zhang2019GANet}, was finally chosen for our framework. 
In comparison to previous works on deformable tissue tracking, which require a substantial amount of spatial and temporal filtering on the depth image \cite{song2017dynamicTissue}, \cite{li2019super}, our method employs a deep stereo matching network for accurate and dense depth estimation, which requires no post-processing on the depth image.

\section{Experiments}
\label{section:experiment}
We evaluated the proposed framework on three open-source datasets for multiple tasks addressing the performance of the surgical tool tracking and deformable tissue tracking. We compared it with the state-of-the-art methods for analysis. 
The experiments were conducted on two identical computers, each containing an Intel\textregistered{} Core\texttrademark{} i9-7940X Processor and NVIDIA's GeForce RTX 2080.

\begin{figure*}[t!]
\centering
\vspace{2mm}
\begin{subfigure}{0.32\textwidth}
\includegraphics[width=1\textwidth]{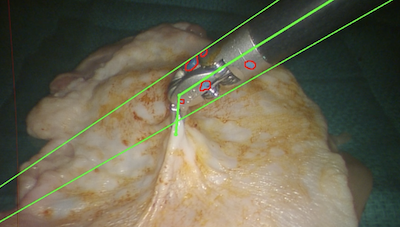}
\vspace{-0.12in}
\end{subfigure}
\begin{subfigure}{0.32\textwidth}
\includegraphics[width=1\textwidth]{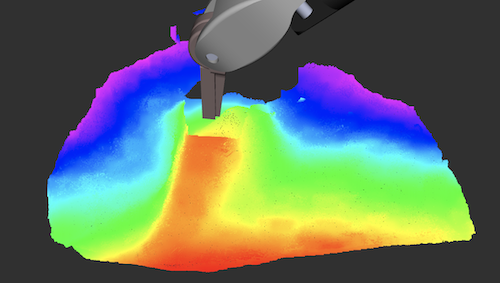}
\vspace{-0.12in}
\end{subfigure}
\begin{subfigure}{0.32\textwidth}
\includegraphics[width=1\textwidth]{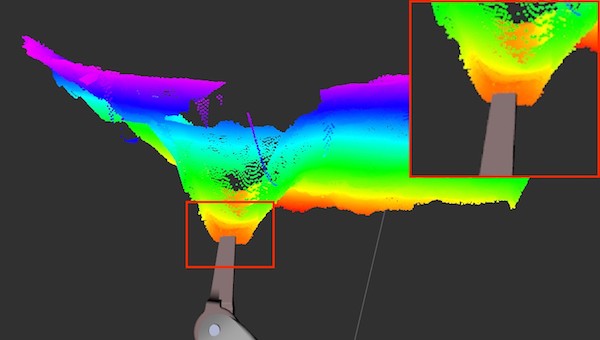}
\vspace{-0.12in}
\end{subfigure}
\begin{subfigure}{0.32\textwidth}
\includegraphics[width=1\textwidth]{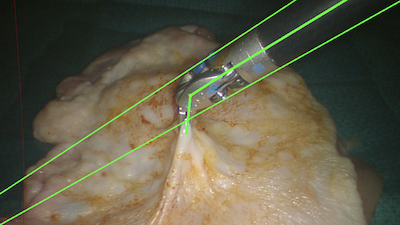}
\vspace{-0.12in}
\end{subfigure}
\begin{subfigure}{0.32\textwidth}
\includegraphics[width=1\textwidth]{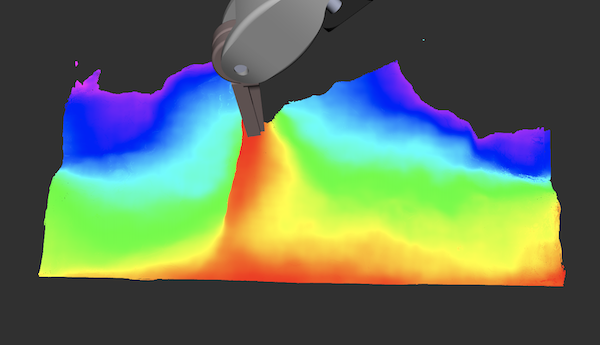}
\vspace{-0.12in}
\end{subfigure}
\begin{subfigure}{0.32\textwidth}
\includegraphics[width=1\textwidth]{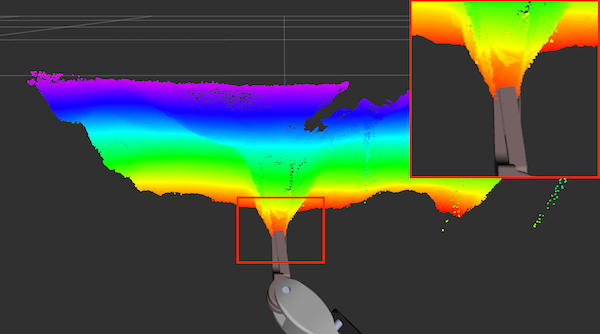}
\vspace{-0.12in}
\end{subfigure}
\caption{Qualitative results from real-time environmental mapping on the SuPer dataset. Top row figures are results from the SuPer framework. Bottom row figures are results from the proposed method, SuPer Deep. From left to right, the figures show: real scene with overlaid tool pose estimation, visualization of deformation reconstruction on RViz with different views. Green lines on the left-most figures indicate the skeleton of the estimated tool pose.}
\label{fig:rviz_results}
\end{figure*}

\subsection{Datasets and Evaluation Metrics}

The Surgical Perception (SuPer)  dataset\footnote{https://sites.google.com/ucsd.edu/super-framework/home} mentioned in \cite{li2019super} is a recording of a repeated tissue manipulation experiment using the da Vinci Research Kit (dVRK) \cite{dvrk}.
The dataset consists of a raw stereo endoscopic video stream and encoder readings from the surgical robot with ground-truth labels for the tool tracking tasks, which consists 50 hand-labeled surgical tool masks.
The tool tracking performance was evaluated by calculating the Intersection-Over-Union (IoU, Jaccard Index) for the rendered tool masks, which are based on estimated tool poses.

The Hamlyn Centre Video Dataset \cite{hamlyn_dataset_1} was used to evaluate the performance of deformable tissue tracking. It includes two video sequences of silicone heart phantom deforming with cardiac motion and consists of ex-vivo endoscopic stereo videos (resolution: 360$\times$288) with depth information generated from CT scans.
The re-projected depth maps of the reconstructed tissue model are evaluated, which is the projection of the entire reconstructed point cloud to the image plane, with each pixel containing a depth value. We calculated the per-pixel root-mean-square (RMS) error of the depth map for every image:
\begin{equation}
    \sqrt{\frac{1}{N_p} \sum_{i,j} (\hat{d}_{i,j} - d_{i,j})^2}
\end{equation}
where $i,j$ is the pixel position, $\hat{d}$ is the estimated depth value, $d$ is the ground truth depth value, and $N_p$ is the total number of pixels for each image. We also reported the percentage of the valid (non-zero) pixels of the depth map.

The da Vinci tool tracking dataset used in \cite{EKF} consists of a stereo video stream and the corresponding kinematic information of the da Vinci\textregistered{} surgical robot. The dataset is used to evaluate surgical tool feature detection and pose estimation. Note that the SuPer dataset has painted markers, and hence this additional experiment ensures that surgical tool feature detector learns surgical tool point features and is not dependent on colored markers. The performance of feature detection is evaluated by calculating the $L_2$ norm of the error in pixels, for the $i$-th feature:
\begin{equation}
    \frac{1}{N} \sum_{n=1}^{N} || \mathbf{h}^i_n - \mathbf{t}^i_n ||_2
\end{equation}
where $N$ is the total number of test images, $\mathbf{h}^i_n$ is the predicted feature point location, and $\mathbf{t}^i_n$ is the ground truth feature point location in the $n$-th image. 
We experiment with varying amounts of hand-labeled training data to illustrate the data efficiency of the proposed surgical tool feature detection method..
Due to lack of ground-truth data for pose estimation, we only provide qualitative results for surgical robotic tool tracking on this dataset.

\subsection{Implementation Details}

\begin{figure*}[t]
    \centering
    \includegraphics[width=1.0\linewidth]{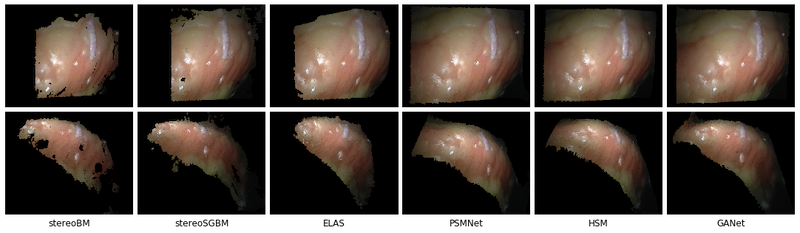}
    \caption{A comparison of tissue reconstruction by fusing the depth maps from different stereo matching algorithms. The first row shows the front view and the second row shows the side view of the reconstruction. Fusing the depth maps generated from DNNs results in more complete and smooth reconstructions.}
    \label{fig:depth_map}
    \vspace{-0.1in}
\end{figure*}

\begin{figure}[t!]
\vspace{2mm}
\begin{subfigure}{0.24\textwidth}
\includegraphics[width=1\textwidth,height=1.1in]{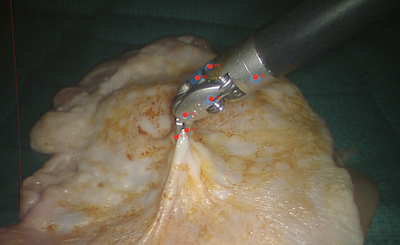}
\vspace{-0.12in}
\end{subfigure}
\begin{subfigure}{0.24\textwidth}
\includegraphics[width=1\textwidth,height=1.1in]{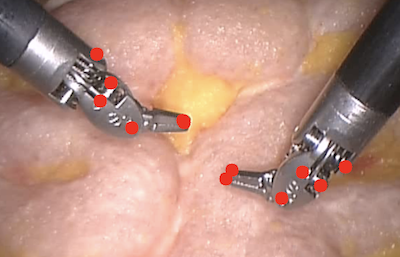}
\vspace{-0.12in}
\end{subfigure}
\\
\begin{subfigure}{0.24\textwidth}
\includegraphics[width=1\textwidth,height=1.1in]{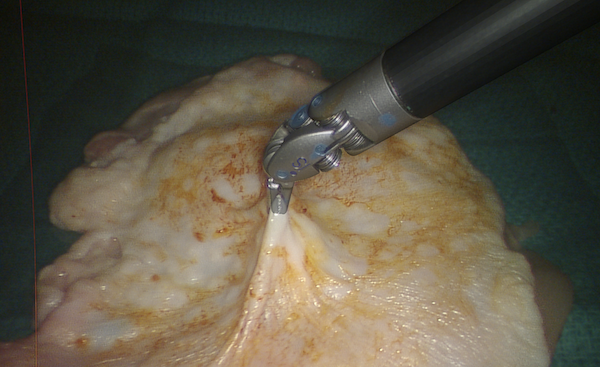}
\vspace{-0.12in}
\end{subfigure}
\begin{subfigure}{0.24\textwidth}
\includegraphics[width=1\textwidth,height=1.1in]{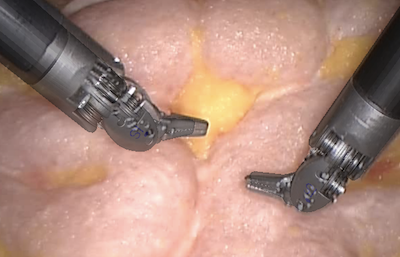}
\vspace{-0.12in}
\end{subfigure}
\caption{Qualitative results of surgical tool tracking. The top row shows the DeepLabCut prediction overlaid on the real images. The bottom row shows an Augmented Reality rendering of the surgical tool \cite{florian_AR} on top of the real images. The renderings are best viewed in color due to near-perfect overlap.}
\label{fig:tool_tracking_result}
\vspace{-0.1in}
\end{figure}

\subsubsection{Surgical Tool Tracking}

For the SuPer dataset, the images were downsampled by 2 before passing to DeepLabCut for feature detection. The weights of DeepLabCut were pre-trained on ImageNet and fine-tuned by training on only 50 hand-labeled images for 7100 iterations with stochastic gradient descent of batch size 1 and learning rate $lr = 0.005$.
For the particle filter, we used 1000 particles with bootstrap approximation on the prediction step and stratified resampling when the number of effective particles dropped below 500. For initialization, the parameterized Lumped Error is set to $[\boldsymbol{\omega}_{0|0}, \mathbf{b}_{0|0}]^T = [0,0,0,0,0,0]^T$ with $\mathbf{\Sigma_{0}} = diag(0.005, 0.005, 0.005, 0.025, 0.025, 0.025)$. For the motion model, the covariance is $\mathbf{\Sigma_{\boldsymbol{\omega},b}} = 0.1*\mathbf{\Sigma_{0}}$, and the $\gamma$ is set to 0.1 for the observation model.
 
\subsubsection{Deformable Tissue Tracking}
For depth map estimation, the raw stereo images were rectified, undistorted, and resized to (640, 480) before being passed into the stereo matching algorithm. Due to the lack of task-specific datasets for surgical environments, the pretrained weights of GA-Net were utilized, which was trained on the Scene Flow dataset from scratch for 10 epochs and fine-tuned on the KITTI2015 dataset for 640 epochs. 
The maximum disparity $D_{max}$ was set to 192. After inverting the disparity, the resulting depth map was fused into the tissue model after subtracting the rendered tool mask, which is dilated by 5 pixels.

For the comparative study, we implemented 3 deep-learning-based and 3 non-deep learning stereo matching algorithms for deformable tissue tracking.
The stereoBM\footnote{{https://docs.opencv.org/4.2.0/d9/dba/classcv\_1\_1StereoBM.html}} and stereoSGBM \cite{SGBM} algorithms are implemented using OpenCV package. ELAS, which is utilized by \cite{song2017dynamicTissue} and \cite{li2019super}, is implemented using the open-source library \cite{stereo_matching}. 
The Pyramid Stereo Matching Network (PSMNet) \cite{Chang_2018} and the Hierarchical Stereo Matching network (HSM) \cite{yang2019hsm} were implemented using the off-the-shelf pretrained weights provided by their original implementations without any task-specific fine-tuning. 

\section{Results}

Qualitative results of the environment mapping on the SuPer dataset are presented in Fig. \ref{fig:rviz_results}. As highlighted in the figures, SuPer Deep provides a larger field of view of the unstructured environment while preserving better details on the reconstruction. 
In comparison to the SuPer framework, more detailed information is captured due to the lack of filtering and smoothing applied on the stereo reconstruction process, which was required in previous implementations of tissue tracking \cite{li2019super}.

\begin{table*}[t]
\vspace{2mm}
    \centering
    \setlength\tabcolsep{0.5em}
    \scalebox{1.2}{
    \begin{tabular}{|c|c|c|c|c|}
        \hline
        \multirow{2}{*}{Method} & \multicolumn{2}{c|}{ Video 1} & \multicolumn{2}{c|}{ Video 2}\\
        \cline{2-5}
         & RMSE & Perc. valid & RMSE & Perc. valid\\
         \hline
        stereoBM + deformable tissue tracker& 23.24 $\pm$ 2.18  & 0.565 $\pm$ 0.037 & 34.02 $\pm$ 2.08 & 0.523 $\pm$ 0.033\\
        stereoSGBM + deformable tissue tracker& 16.84 $\pm$ 1.99 & 0.713 $\pm$ 0.038 &
        24.68 $\pm$ 1.84 & 0.683 $\pm$ 0.032\\
        ELAS + deformable tissue tracker & 16.12 $\pm$ 2.22 & 0.716 $\pm$ 0.044& 22.03 $\pm$ 2.89 & 0.719 $\pm$ 0.049\\
        PSMNet + deformable tissue tracker& 5.64 $\pm$ 1.48 & 0.940 $\pm$ 0.023 & 8.23 $\pm$ 1.32 & 0.939 $\pm$ 0.014   \\
        HSM + deformable tissue tracker& 5.33 $\pm$ 1.36& 0.938 $\pm$ 0.023& 6.73 $\pm$ 1.28 & 0.946 $\pm$ 0.020
        \\
        GA-Net + deformable tissue tracker & \textbf{4.87 $\pm$ 1.55} & \textbf{0.947 $\pm$ 0.026} & \textbf{6.20 $\pm$ 1.57} & \textbf{0.957 $\pm$ 0.024}   \\
        \hline
    \end{tabular}}
    \caption{A comparative study of the deformable tissue tracking performance for the Hamlyn Centre heart phantom dataset. Depth maps obtained from different stereo matching algorithms are fused by the deformable tissue tracker to update the tissue model, and the re-projection error of reconstructed deformable tissue surface are calculated. The percentage of valid pixels and per-pixel RMS error are measured. }
    \vspace{-0.15in}
    \label{table:average_rms}

\end{table*}

\subsubsection{Deformable Tissue Tracking}

Using the Hamyln Centre Video Dataset, the deformable tissue tracking results were compared by combining popular stereo matching algorithms with the deformable tissue tracker.
We visualize the reconstruction results by fusing the first 10 estimated depth maps from each algorithm in Fig. \ref{fig:depth_map}. Deep-learning-based algorithms generally provide dense and consistent matches and result in realistic tissue reconstructions.
We calculated the average per-pixel RMS error on the re-projected depth maps, and the quantitative results are shown in Table \ref{table:average_rms}. 
Deep-learning-based approaches achieve much lower per-pixel RMS error with the higher percentage of the valid pixel, which confirms the observations from the environment mapping results in Fig. \ref{fig:rviz_results} and Fig. \ref{fig:depth_map}.

%
%


\subsubsection{Surgical Tool Tracking}
Fig. \ref{fig:dlc_exp} shows the feature detection performance of the DeepLabCut with varying numbers of training samples. By leveraging transfer learning, the feature detector is able to achieve high performance on detecting surgical tool features using few training samples.

For the tool tracking task, SuPer Deep achieved $\mathbf{91.0\%}$ mean IoU on the SuPer tool segmentation task, which is a significant improvement on the original method (SuPer: 82.8$\%$). Notably, SuPer Deep does marker-less tool tracking while the former utilizes painted markers. Qualitative results of the tool tracking are presented in Fig. \ref{fig:tool_tracking_result}, where we experimented with our tool tracker on both the SuPer dataset and the da Vinci tool tracking dataset. In the visualization, the Augmented Reality rendering from the estimated tool pose produces a near-perfect overlap with the tool on real images.

\section{Discussion}

\begin{figure}[t]
\vspace{2mm}
    \centering
    \includegraphics[width=1.0\linewidth]{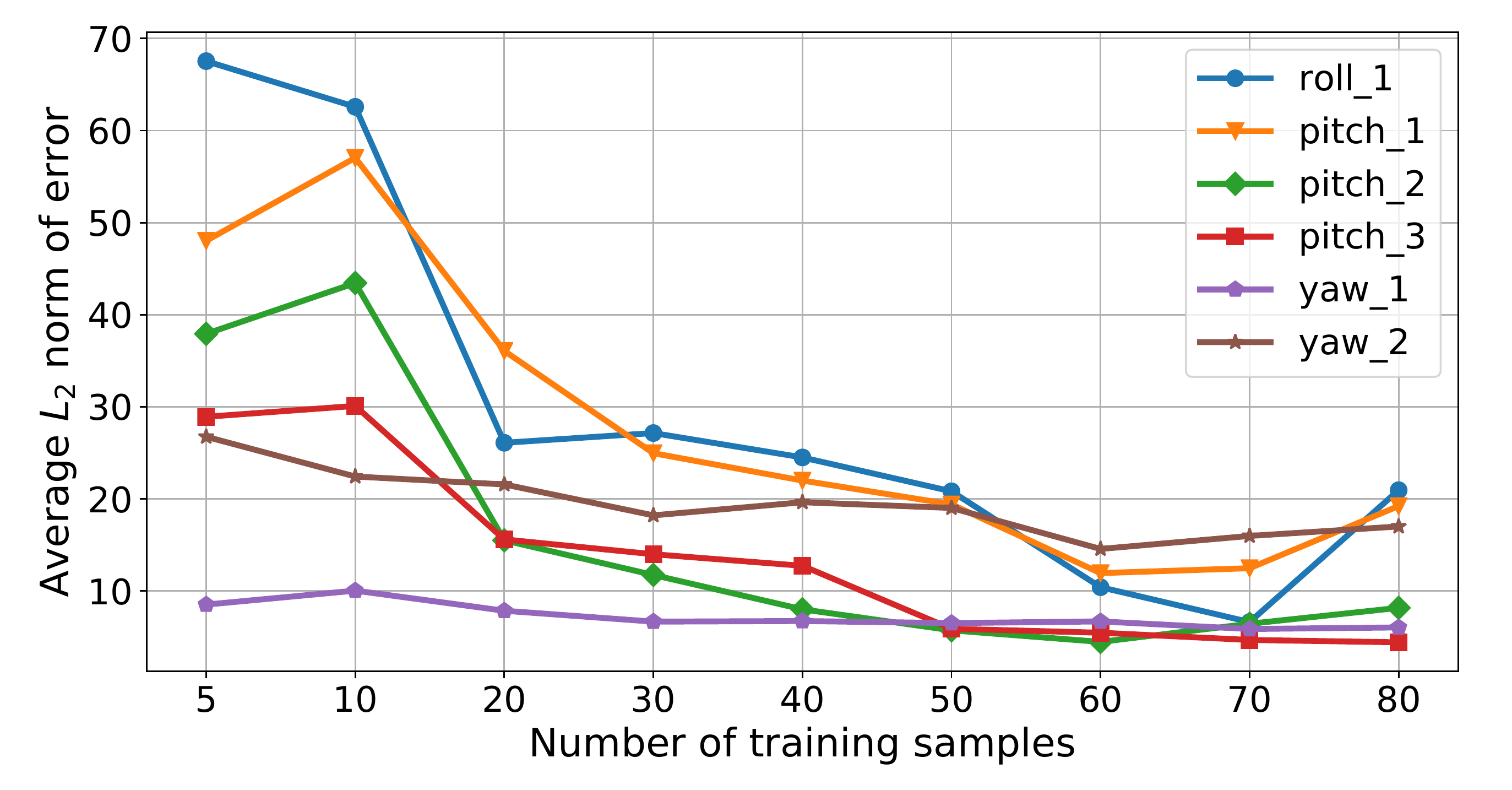}
    \caption{Plot of average detection error of features on surgical tool against the amount of training data. As is evident by the plot, only 60 to 80 training samples are required to train an accurate model.}
    \label{fig:dlc_exp}
    \vspace{-0.15in}
\end{figure}

The experimental results show SuPer Deep achieves excellent performances in both surgical tool tracking and deformable tissue tracking. 
By utilizing deep neural networks, SuPer Deep produces more consistent depth maps and achieves accurate tool pose estimation. The latter also helps to reduce the dilation of the tool mask, which reduces the amount of information lost. As the visualizations shows in Fig. \ref{fig:rviz_results}, SuPer Deep's reconstruction has the surgical tool \textit{touching} the point cloud (as opposed to just being above the point cloud).

\begin{figure}[t!]
\centering
\vspace{2mm}
\begin{subfigure}{0.23\textwidth}
\includegraphics[width=1\textwidth]{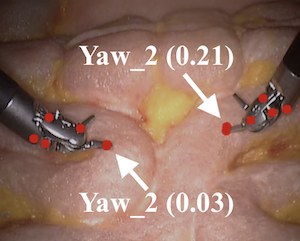}
\vspace{-0.12in}
\end{subfigure}
\begin{subfigure}{0.23\textwidth}
\includegraphics[width=1\textwidth]{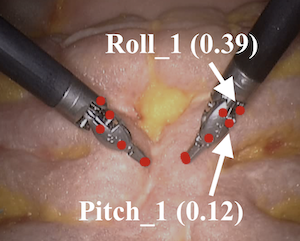}
\vspace{-0.12in}
\end{subfigure}
\caption{Failure cases in tool point feature detection. In parenthesis are  the corresponding confidence scores. 
Note that when a failure occurs the DNN outputs low confidence hence making the missed detection not as detrimental to our tracking algorithm.}
\label{fig:dlc_fail}
\vspace{-0.1in}
\end{figure}

There are occasional failures in feature detection on the surgical tool, owing mainly to the symmetry of the tool parts, for example, the \textit{Roll\_1}, \textit{Pitch\_1} and \textit{Yaw\_2} features. As shown in Fig. \ref{fig:dlc_exp}, detecting those features is more challenging compared to other ones. The misdetections are, however, of low confidence. Hence they are handled by the probability weighting of the detected points in the observation model of the particle filter. In Fig. \ref{fig:dlc_fail}, for instance, one of the grippers of the tools is misdetected, but has substantially lower confidence; meanwhile the correctly detected points have confidence scores higher than 70\%. Similarly, in the second case, two features are detected on the wrong side of the shaft due to the surgical tools symmetry, but again with low confidence and hence is not detrimental to the pose estimation. Overall, the feature detection is robust and results in accurate perception.

\section{Conclusion}

Deep learning has not been utilized as a major tool in surgical robotic perception, with a lack of training data as the primary bottleneck \cite{li2019super}. The SuPer Deep framework, incorporating two deep neural networks as major components, shows that the challenge of insufficient data is surmountable. Using transfer learning, even on limited training data, the framework  accomplishes excellent feature detection for surgical scene perception. Our comparative study on deformable tissue tracking, which utilizes the deep neural networks with only pretrained weights, shows that deep learning techniques can be applied for stereo reconstruction and gives a performance evaluation on these techniques applied to a surgical context.

Currently, we believe that the major limitation of the SuPer Deep framework is its high computation power. 
Running multiple deep neural networks in real-time requires multiple processing units, which limits the update rates of the trackers. 
Lightweight deep neural networks will be ideal for real-time surgical applications, if adapted without compromising on accuracy. 
As recent progress has been made on deep-learning-based reconstruction and rendering techniques \cite{jiang2020sdfdiff}, \cite{park2019latentfusion}, a future direction could be utilizing a learnable tissue tracker and tool tracker to further optimize the perception framework.
Another direction to pursue is surgical task automation. By using the perceived environment as feedback, controllers applied to the surgical tool will be able to accomplish tasks in unstructured, deforming surgical environments. 

\balance
\bibliographystyle{ieeetr}
\bibliography{references}

\end{document}